\documentclass[journal]{IEEEtran}

\usepackage{balance}
\usepackage{times}
\usepackage{epsfig}
\usepackage{graphicx}
\usepackage{amsmath}
\usepackage{amssymb}
\usepackage{multirow}
\usepackage{amsmath}
\usepackage{xcolor}
\usepackage{caption}
\usepackage[linesnumbered,ruled]{algorithm2e}
\usepackage{microtype}
\usepackage{grfext}

\def\ie{{\em i.e.}}
\def\eg{{\em e.g.}}
\def\etal{{\em et al.}}

\newcommand{\figref}[1]{Fig. \ref{#1}}
\newcommand{\tabref}[1]{Tab. \ref{#1}}

\newcommand{\secref}[1]{Section \ref{#1}}

\newcommand{\mc}[1]{\mathcal{#1}}
\newcommand{\mb}[1]{\mathbb{#1}}

\newcommand{\bs}[1]{\boldsymbol{\texttt{#1}}}

\usepackage{epstopdf}
\usepackage{ragged2e}
\usepackage{booktabs}
\usepackage{color}
\usepackage{multirow}
\usepackage{bm}
\usepackage{bbm}
\usepackage[misc]{ifsym}
\usepackage{graphicx}
\usepackage{amsmath}
\usepackage{amsfonts}
\usepackage{enumerate}

\usepackage{pifont}
\usepackage[colorlinks,linkcolor=black]{hyperref}
\usepackage{mathrsfs}

\begin{document}
\title{Towards Unified Co-Speech Gesture Generation via Hierarchical Implicit Periodicity Learning}

\author{Xin~Guo, Yifan~Zhao,~\IEEEmembership{Member,~IEEE},~Jia~Li,~\IEEEmembership{Senior Member,~IEEE} 
\IEEEcompsocitemizethanks{
\IEEEcompsocthanksitem X. Guo, Y. Zhao, and J. Li are with the State Key Laboratory of Virtual Reality Technology and Systems, School of Computer Science and Engineering \textnormal{\&}Qingdao Research Institute, Beihang University, Beijing, 100191, China.
\IEEEcompsocthanksitem Y. Zhao and J. Li are the corresponding authors.
(E-mail: zhaoyf@buaa.edu.cn, jiali@buaa.edu.cn).
}}

\maketitle

\begin{figure*}[t]
\begin{center}
   \includegraphics[width=1.0\linewidth]{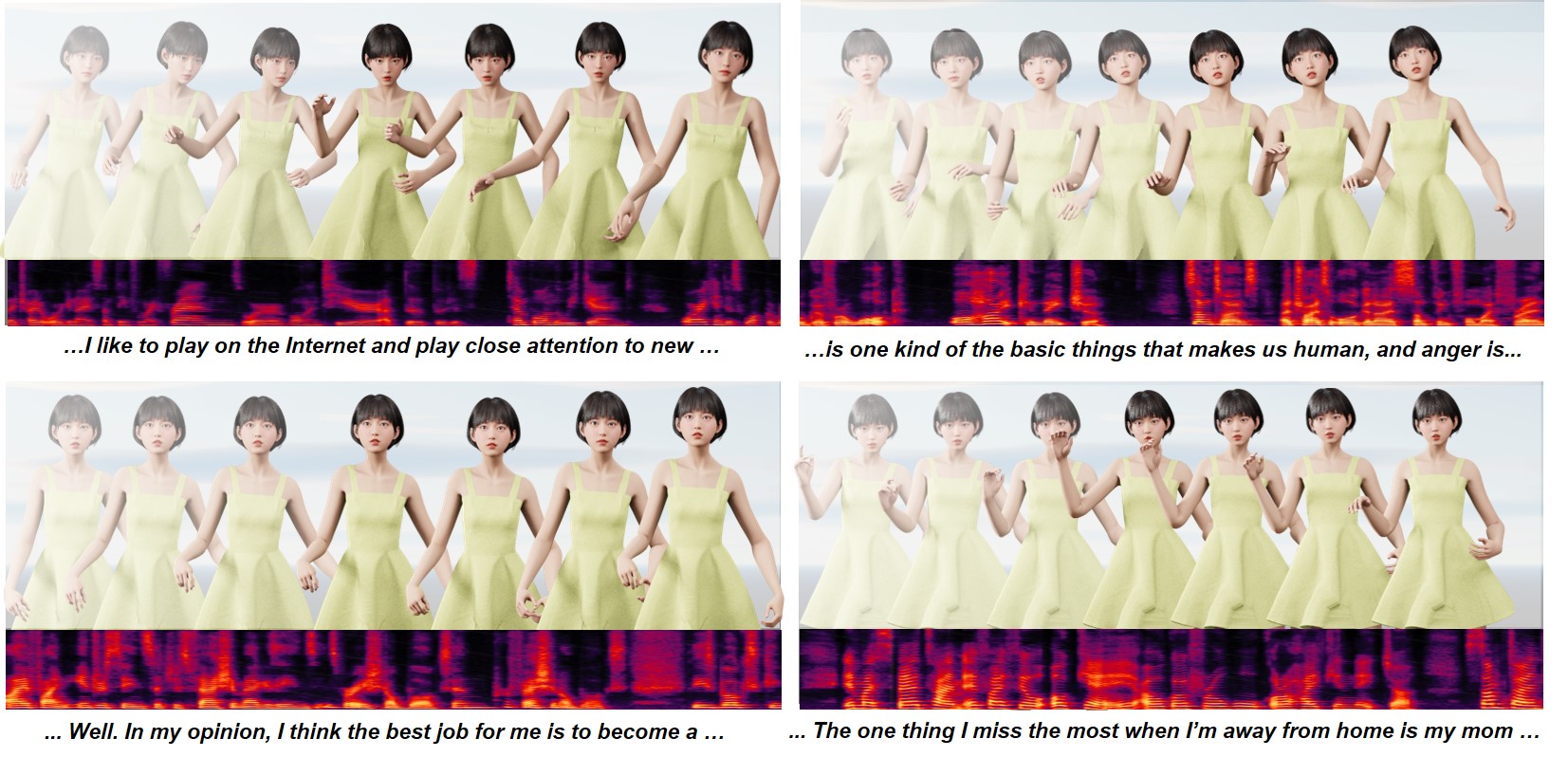}
\end{center}
   \caption{Human gesture with 3D avatars synthesized by our proposed pipeline on different inputs. From top to bottom are the generated character animation, input audio, and text.}
\label{fig:motivation}
\end{figure*}

\begin{abstract}
\justifying Generating 3D-based body movements from speech shows great potential in extensive downstream applications, while it still suffers challenges in imitating realistic human movements. Predominant research efforts focus on end-to-end generation schemes to generate co-speech gestures, spanning GANs, VQ-VAE, and recent diffusion models. As an ill-posed problem, in this paper, we argue that these prevailing learning schemes fail to model crucial inter- and intra-correlations across different motion units,~\ie head, body, and hands, thus leading to unnatural movements and poor coordination. To delve into these intrinsic correlations, we propose a unified Hierarchical Implicit Periodicity (\textbf{HIP}) learning approach for audio-inspired 3D gesture generation. Different from predominant research, our approach models this multi-modal implicit relationship by two explicit technique insights: i) To disentangle the complicated gesture movements, we first explore the gesture motion phase manifolds with periodic autoencoders to imitate human natures from realistic distributions while incorporating non-period ones from current latent states for instance-level diversities. ii) To model the hierarchical relationship of face motions, body gestures, and hand movements, driving the animation with cascaded guidance during learning. We exhibit our proposed approach on 3D avatars and extensive experiments show our method outperforms the state-of-the-art co-speech gesture generation methods by both quantitative and qualitative evaluations.
Code and models will be publicly available.
\end{abstract}

\begin{IEEEkeywords}
3D-based body movements, Hierarchical implicit periodicity, Phase manifolds, Multi-modal implicit relationship
\end{IEEEkeywords}

\IEEEpeerreviewmaketitle

\section{Introduction}\label{sec:intro}

When a person attempts to articulate his thoughts, two distinct modalities are employed: speech and physical gestures. Verbal communication serves as the principal means for expressing one's ideas, while gestures offer a complementary way to concretize content and emotional expressions, thereby enhancing the comprehensibility of the conveyed message to others. For instance, during greetings or interpersonal interactions, people employ a repertoire of gestures alongside their verbal communication, with these gestures indirectly revealing their emotional disposition towards the interlocutor. Facial expressions and body gestures also exhibit a degree of coordination, such as the discernible divergence in gestures when individuals experience varying degrees of happiness. Consequently, the exploration of speech-driven human body gesture generation has emerged as a promising research avenue.

Co-speech gesture generation, an ill-posed one-to-many mapping problem, requires modeling both intra-unit (within face, body, hands) and inter-unit correlations for coherent gesture production. The methods include: retrieval-based~\cite{lee2002interactive,arikan2002interactive,kovar2023motion,yang2023qpgesture} which decomposes gestures into action units, extracts conditional features, and retrieves similar motions from databases, achieving high controllability but limited to database content. End-to-end models~\cite{liu2022beat,ghorbani2023zeroeggs,zhu2023taming,zhi2023livelyspeaker} use architectures like RNNs and Transformers to directly map audio to gestures, supporting complex cross-modal mappings and generating diverse gestures. Two-stage methods~\cite{xu2024mambatalk,yi2023generating} map audio to intermediate latent codes before decoding them into motion sequences. Although they enhance gesture diversity, both end-to-end and two-stage approaches often treat body movements as a single aggregated signal, overlooking structured inter and intra-correlations, leading to unstable and semantically misaligned gesture generation. Hierarchical methods~\cite{yi2023generating,habibie2021learning,liu2023emage,chen2024diffsheg,liu2022learning} attempt to model different body parts separately. However, they still lack a clear correlation hierarchy: Habibie~\cite{habibie2021learning} and TALKSHOW~\cite{yi2023generating} independently generate facial and body motions without cross-part coordination; EMAGE~\cite{liu2023emage} predicts masked body parts simultaneously to allow mutual influence but lacks a dependency order; DiffSHEG \cite{chen2024diffsheg} first generates facial motions and then predicts the entire body as a single signal, ignoring structured relationships within the body itself; and HA2G \cite{liu2022learning} generates body parts in a stepwise manner but omits facial information entirely.

While existing methods have advanced co-speech gesture generation, they often neglect the dependencies among different body parts. Human motions follow certain morphological and physical rules, and the movement of one human part often influences the motion of other parts. Gestures follow a natural hierarchy: facial expressions are most strongly tied to speech content and emotion, body motions are shaped by facial emotions and speaking state, and hand gestures depend on body dynamics and semantic context. However, existing hierarchical approaches have not effectively integrated this progressive mechanism from strongly associated units to weakly associated units. In addition, the distinction between periodic and non-periodic motion is often overlooked. Periodic gestures, such as nodding and smiling, exhibit stable rhythms and regularity, whereas non-periodic gestures, such as emphatic beats or unique expressions, introduce variation and highlight semantic focus. Thus here arise two natural questions: \textcircled{1} how to model the temporally physical intra-correlations within each unit? \textcircled{2} how to model the inter-correlations across different body moving units (\eg, face, body, and hand)?

Keeping question \textcircled{1} in mind, co-speech human motions follow certain intrinsic and basic rules that are seriously omitted by the prevailing end-to-end models. To explore this, we start with an empirical analysis of implicit human motion rules from a new perspective of physical periodicity. To model these physical rules, inspired by periodicity learning \cite{starke2022deepphase}, we make attempts to model the phase manifold of co-speech gestures in this task, including body and hand movements. However, this intuitive manner leads to a severe loss of individual diversities,~\ie, generating repetitive movements.  The reason for this phenomenon is that unlike activities with strong periodicity such as walking or running, gestures encompass a substantial amount of non-periodic movements. Toward this end, we disentangle these complicated gesture movements as two terms, periodicity for common characteristics and non-periodicity for instance-level diversities. Based on this finding, we develop a periodicity disentanglement module to extract the common periodic phase from realistic training data while incorporating the instance-level latent features to enhance the non-periodic diversities. Our motivation is to enhance the naturalness of generated gestures by employing physical rules to better capture the periodic components in gesture movements, driving the network to learn more effectively.

For question \textcircled{2}, we propose a unified hierarchical attribute guidance module to model the correlations of multiple moving units. As the head units include the strongest relations with the speech audio,~\eg, lip movements, and facial emotions, we adopt the head units as the predominant learning guidance for generating body movements, while the hand gestures perform a subordinate relationship with body gestures. Different from the previous work \cite{liu2022beat} that utilized facial capture data as input information into the model, we establish an audio-to-face prediction model to extract facial features. With these key insights for \textcircled{1} and \textcircled{2}, we develop a unified learning framework that incorporates individual IDs and emotional labels, controlling the network to perform diverse generations for different scenarios. To summarize, the main contributions of our work are as follows:

\begin{enumerate}
    \item We start from a novel view to disentangle the complicated gesture movements and propose a periodicity disentanglement module to jointly model the common motion rules while incorporating instance-level diversities.
    \item We propose a hierarchical attribute guidance module to model the correlations of multiple moving units, enhancing strong correlations while preserving the subordinate weak correlations.
    \item We develop a unified co-speech gesture learning framework with multi-modal inputs and experimental results demonstrate that our proposed framework outperforms the existing state-of-the-art methods in both subjective and objective studies.
\end{enumerate} 

The remainder of this paper is organized as follows: ~\secref{sec:relatedwork} describes the related works and ~\secref{sec:method} presents the proposed hierarchical implicit periodicity learning for co-speech gesture generation. Qualitative and quantitative experimental results are presented in~\secref{sec:exp} and~\secref{sec:conclusion} finally concludes this paper.

\section{Related Works}\label{sec:relatedwork}

\noindent\textbf{Audio2Face Generation.} Due to the strong correlation between audio and the head, researchers have explored the correspondence between audio and head animations. In the task of generating talking head videos, researchers have conducted a series of works \cite{zakharov2019few,chen2019hierarchical,thies2020neural,mittal2020animating,zhang20213d,guo2021ad,zhang2021facial,ji2021audio,hong2022depth,gan2023efficient} based on speech-driven or image-driven approaches. For example, Hong \etal~ \cite{hong2023implicit} proposed a module that learnt prior knowledge about the appearance and structure of the head from data samples of multiple individuals, and compensated the warped areas during the generation process. Yu \etal~\cite {yu2023talking} adopted Stable Diffusion \cite{ho2020denoising} to explore the mapping relationship between audio and lip-irrelevant facial motions. Different from these methods that focused on 2D video generation, researchers attempted to explore the task of speech-driven 3D face animations synthesis \cite{chen2022transformer,cudeiro2019capture,fan2022faceformer,hussen2020modality,karras2017audio,liu2021geometry,richard2021meshtalk,peng2023emotalk}. Specifically, Fan \etal~\cite {fan2022faceformer} proposed an encoder-decoder model based on transformer, taking the original audio as input, and autonomously generates a sequence of animated 3D facial meshes. Thambiraja \etal
~\cite{thambiraja2023imitator} proposed a model that learnt prior knowledge on a large facial expression dataset to optimize the talking style of the identity-specific person. 

\noindent\textbf{Motion Synthesis.} In the early works \cite{arikan2002interactive, kovar2023motion, lee2002interactive}, researchers built a motion-graph to synthesis the motion sequences. The generated motions were the origin data in the graph and the workers defined a distance metric to select the next node in the graph based on the previous data. With development of deep learning methods, researchers adopted Feedforward \cite{holden2017phase, starke2019neural,starke2020local,starke2022deepphase}, RNN \cite{lee2018interactive, harvey2020robust}, GAN \cite{henter2020moglow, ling2020character, valle2021transflower} and RL \cite{peng2016terrain, peng2017deeploco, peng2018deepmimic, cho2021motion, lee2021learning} to generate the motion sequences. Holden \etal~\cite{holden2017phase} proposed Phase-functioned Neural Networks that define periodic variables based on foot contact with the ground. But it only supported idle walking and running. Based on this approach, Starke \etal~\cite{starke2019neural} introduced the Neural State Machine, which defined a global phase label for complex motions such as locomotion, sitting, standing, lifting, and collision avoidance. Furthermore, Starke \etal~\cite{starke2020local} extended this concept to play basketball which extracted the local phase of each limb to predict the motion of the next frame. Mason \etal~\cite{mason2022real} adopted transformer \cite{vaswani2017attention} and local motion phases to model the motion content and style modulation. Recently, Starke \etal~\cite{starke2022deepphase} combined FFT with neural networks to predict the periodic parameters of movements in an unsupervised manner. However, this method performed poorly in actions that included non-periodic components, such as body gestures accompanying speech. In our work, for non-periodic movements, we incorporate a non-periodic branch based on this approach to address this issue.

\noindent\textbf{Co-Speech Gesture Generation.} Driving avatar gestures is a task with a wide range of applications. The early researches \cite{cassell1994animated,cassell2001beat,kipp2005gesture,kopp2006towards,huang2012robot,marsella2013virtual} often defined a rule-based method to mapping speech units to gesture fragments. The advantages of the rule-based model were easy to produce controllable results, but it was labor-intensive. To solve this problem, researchers adopted statistical models \cite{neff2008gesture,levine2009real,levine2010gesture} to learn the mapping rules from speech units to gesture clips. Recent data-driven approaches adopted CNN \cite{habibie2021learning}, RNN \cite{yoon2019robots,yoon2020speech,bhattacharya2021speech2affectivegestures,liu2022learning,liu2022beat}, VAEs \cite{li2021audio2gestures,ghorbani2023zeroeggs}, VQ-VAE ~\cite{liu2022audio,yazdian2022gesture2vec,yi2023generating,liu2024towards,xu2024mambatalk,liu2023emage}, Transformers~\cite{bhattacharya2021text2gestures} and Stable Diffusion~\cite{ao2023gesturediffuclip,zhu2023taming,zhi2023livelyspeaker,ahuja2023continual,yang2023diffusestylegesture,yang2023unifiedgesture,zhu2023taming,yang2023diffusestylegesture,chen2024diffsheg} to learn the relationship between speech and gestures. For example, Ao \etal~\cite{ao2023gesturediffuclip} introduced a novel framework that utilized the CLIP~\cite{radford2021learning} and VQ-VAE~\cite{van2017neural} to explore the potential relationship between gesture and transcript, then adopted Stable Diffusion~\cite{rombach2022high} to decode the audio, transcript and style features to the target motion sequences. However, these end-to-end methods do not consider the physical rules underlying gesture synthesis tasks to assist in the generation of actions.

In terms of generating full-body animations, Habibie \etal~\cite{habibie2021learning} first proposed a method to generate facial and body animations simultaneously. Yi \etal~\cite{yi2023generating} introduced TALKSHOW that quantified the body and hand respectively. However, these methods did not consider the synergy between facial and body gestures. DiffSHEG \cite{chen2024diffsheg} has made some progress in this area, but it lacks exploration into the principles of body motion. In our research, we establish a multi-level generative framework that maps audio to facial, body, and hand animations, and further synthesize more natural and rhythmically appropriate human motions by incorporating the underlying patterns in gesture movements that align with the speech content and rhythm.

\section{Approach}\label{sec:method}

\subsection{Implicit Correlation: An Empirical Analysis}\label{sec:movitation}

\textbf{Implicit correlations behind human motions.} 
Co-speech human motion contains rich implicit information, including certain temporal characteristics, morphological rules, and multi-modal alignment relationships. Pre-dominant end-to-end frameworks follow a data-driven trend while suffering from inferior generation quality when facing extreme cases and unseen scenarios. These works fail to model the implicit physical rules behind human gesture motions. In this paper, we argue for the discovery of implicit human motions by disentangling this process into periodic and non-periodic features. We thus employ the Fast Fourier Transform (FFT) technique, which has been widely utilized in previous studies to extract periodic components from motion signals, effectively capturing repetitive behaviors such as walking and running. Previous studies \cite{starke2022deepphase, yang2023qpgesture} have also proved the presence and significance of periodic components in human motion. However, non-periodic components, usually transient, task-specific, or environment-driven motion patterns, have largely been overlooked. These non-periodic components contain important information about subtle, non-repetitive motion features, such as gesture details and emotional expressions.

\begin{figure}[t]
\begin{center}
   \includegraphics[width=1.0\linewidth]{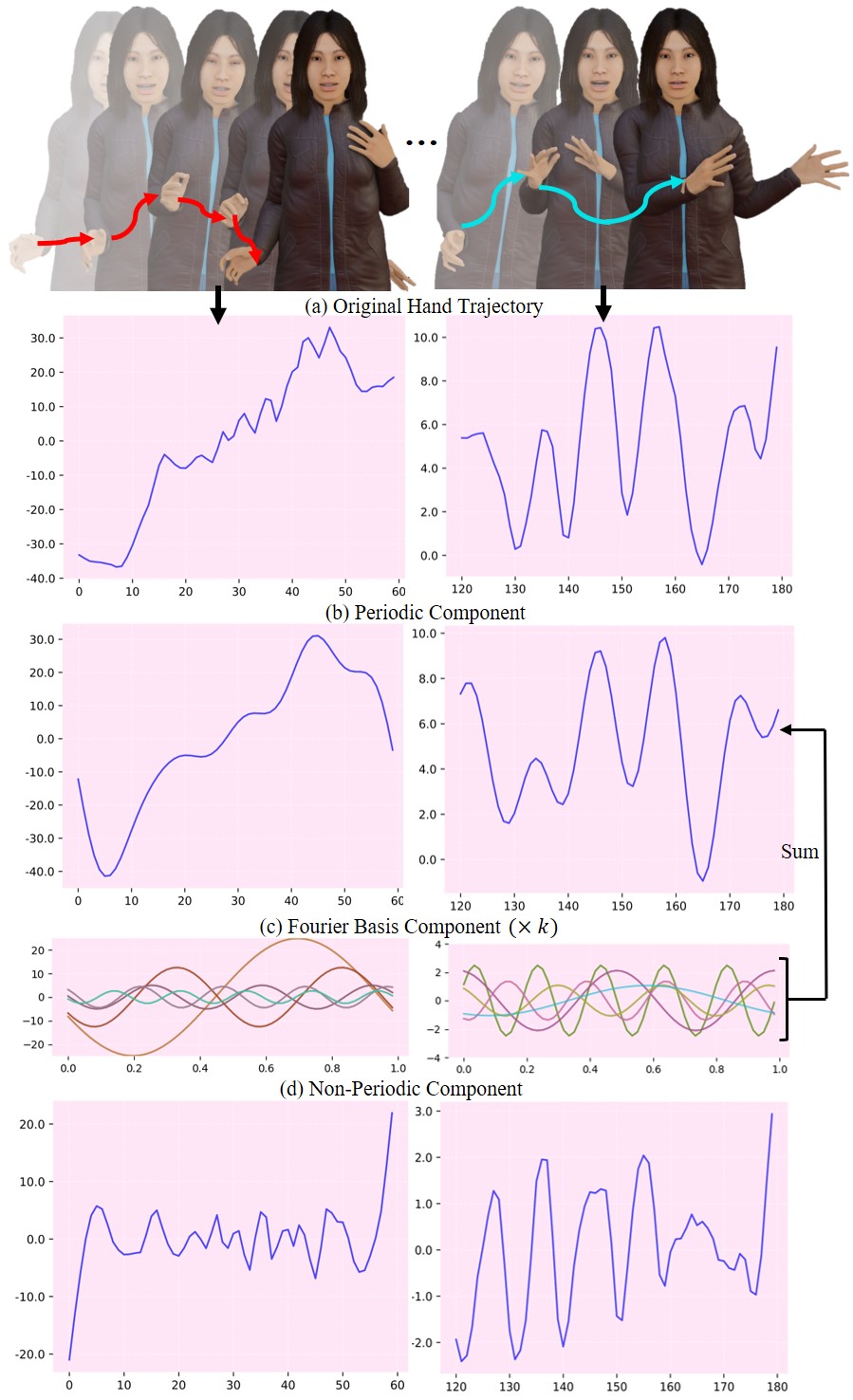}
\end{center}
   \caption{\textbf{Empirical Study}. From top to bottom: the human body motion diagram, the motion trajectory of the hand node in the y-direction, the extracted periodic components, Fourier basis components, and non-periodic components. The periodic components are obtained by summing a limited number of $k$ Fourier basis functions, where different colors in (c) indicate different Fourier basis functions.}
\label{fig:viseuler}
\end{figure}

To illustrate this phenomenon, we construct an empirical study on the widely-used BEAT \cite{liu2022beat} dataset, as shown in~\figref{fig:viseuler}. We visualize the motion trajectory of the hand joint along the Y-axis (\figref{fig:viseuler} (a)) and extract the periodic components (\figref{fig:viseuler} (b)) of the motion by fitting it with $k$ Fourier basis functions (\figref{fig:viseuler} (c)) using Fourier Transform. The difference between the original trajectory and the periodic components represents the non-periodic components (\figref{fig:viseuler} (d)). As shown in the \figref{fig:viseuler}, the hand movement trajectory exhibits two distinct forms: non-periodic (\textcolor{red}{red}) and periodic patterns (\textcolor{cyan!50}{blue}). For both cases, the extracted periodic components reflect the basic motion patterns of the joints, with a value range similar to the original input. The non-periodic components have values distributed around 0, representing the finer details of the movement.

\begin{figure*}[t]
\begin{center}
   \includegraphics[width=1.0\linewidth]{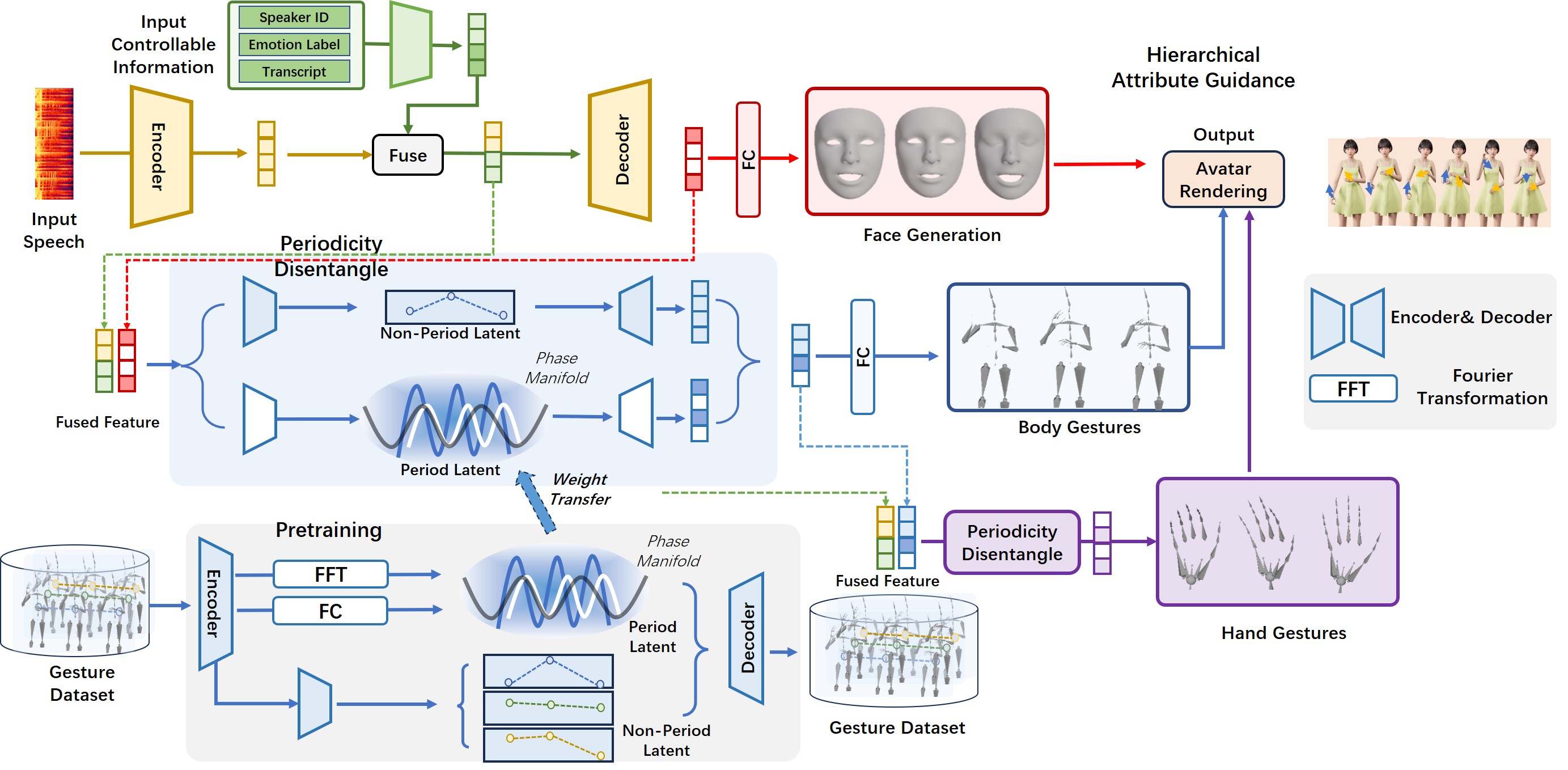}
\end{center}
   \caption{The pipeline of our proposed Hierarchical Implicit Periodicity learning method. Taking the joint speech and controllable information, we first disentangle the gesture movements with period phase latent and non-period individual latent to depict the generated gestures, while the period phase manifold could be pre-learned from the holistic dataset. We then develop a hierarchical attributed guidance to drive the gesture generation in a cascaded manner.  }
\label{fig:pipeline}
\end{figure*}

Based on the above empirical results, here we reach the following observations:
\textcircled{1} Co-speech human motions follow certain topological rules, and the trajectory shows clear \textbf{periodicity};
\textcircled{2} Beyond these periodicities, human motions also present specific characteristics when facing different speech inputs, which we call \textbf{non-periodic} motions.  Based on these observations, we introduce Hierarchical Implicit Periodicity (HIP) learning to model this complex system in speech gesture generation tasks, as shown in \figref{fig:pipeline}. We first introduce periodicity disentanglement (\secref{sec:PD}) to model the regular moving routines for \textcircled{1} while learning the disentangled non-periodicity for \textcircled{2}. To model the inter-correlations of human motions, we then construct the face animation generator for audio consistency lip and facial movements in \secref{sec:face} and then propose the unified hierarchical attribute guidance framework in \secref{sec:hie} to model the implicit correlations among multiple gesture units.

\noindent \textbf{Symbol notations.} Given a piece of speech $\mc{C}_{1:T}^v$, we denote the face, body, hand animations as $\mc{C}^{m} \in \mathbb{R}^{T \times 52}$, $\mc{C}^{body} \in \mathbb{R}^{T \times 27}$, and $\mc{C}^{hand} \in \mathbb{R}^{T \times 114}$, where $T$ is the frames, respectively. $\mc{C}^{bh} \in \mathbb{R}^{T \times 141}$ is defined as a combination of body and hand movements. In addition, as the joint controllable input, the speaker ID, transcript, and emotion label that corresponds to the segment are defined as $\mc{C}^{id}$, $\mc{C}^{text}$ and $\mc{C}^{emo}$.

\subsection{Periodicity Disentanglement}\label{sec:PD}

From the \figref{fig:viseuler} and the analysis in \secref{sec:movitation}, it can be seen that the intra-correlation of the motion contains underlying periodicity. Different from the QPGesture \cite{yang2023qpgesture} which used the angle velocity as the input, we select the velocity of the world coordinate system. To extract the periodic features within gestures, we build the periodicity disentanglement module (PD) and adopt DeepPhase \cite{starke2022deepphase} as the backbone to extract the phase manifold of the motions. Throughout the training process, apart from aiming to reconstruct the input, each feature space within the latent space exists in the form of periodic functions. With this specific design, the goal of the model is to learn the periodic features present within the motions. But different from the actions like walking and running which exhibit strong periodicity, co-speech gestures are weaker than them. Therefore, a non-periodic branch is added to encode the dimensionality-reduced gestures. Through these two distinct branches, the motions are disentangled into periodic and non-periodic features. Given a sequence of gestures $\mc{C}_{1:T}^{bh}$, it is initially input into an encoder $\mc{E}_d$ which is composed of $1D$ convolutions to get a lower-dimensional motion features $\textbf{y}$, which can be formulated as $\textbf{y}=\mc{E}_d(\mc{C}_{1:T}^{bh})$, where $\textbf{y} \in \mb{R}^{T \times \mathcal{N}}$, and $\mathcal{N}$ indicates the number of channels in the following periodic module. Then, the gesture movements are disentangled into periodicity modeling and individual non-periodicity as follows.

\noindent \textbf{Periodicity modeling.} In terms of the period modeling, to extract periodic features of the gestures, each channel of the embedding in the model uses a sinusoidal function to parameterize the extracted motion features $\textbf{y}$. The parameters in the channel include amplitude $\textbf{A}$, frequency $\textbf{F}$, offset $\textbf{B}$, and phase shift $\textbf{S}$. To calculate $\textbf{A},\textbf{F},\textbf{B}\in \mb{R}^{K}$, $K=\frac{T}{2}$, we follow the work \cite{starke2022deepphase} and adopt differentiable real Fast Fourier Transform (FFT) layer to each channel of $\textbf{y}$, which convert the features $\textbf{y}$ of time to the frequency domain $\textbf{Q}$,
\begin{equation}
\begin{split}
  \textbf{Q}_{z,j} = \bs{FFT}(\textbf{y}_z)_j = \sum_{t=0}^{T-1}\textbf{y}_{z,t} \cdot \exp(-i2\pi jt/T).
  \label{eq:mtrain}
\end{split}
\end{equation}
Subsequently, the power spectrum $\textbf{P}$ of each channel is obtained through element-wise operations. Hence the $\textbf{A},\textbf{F},\textbf{B} \in \mb{R}^{K}$ in the $i$-th channel can be calculated by:
\begin{equation}
\begin{split}
  \textbf{A}_i = \sqrt{\frac{2}{T}\sum_{j=1}^{K}\textbf{P}_{i,j}}, 
  \textbf{F}_i = \frac{\sum_{j=1}^{T}\alpha_j \textbf{P}_{i,j}}{\sum_{j=1}^{K}\textbf{P}_{i,j}}, 
  \textbf{B}_i = \frac{\textbf{Q}_{i,0}}{T},
  \label{eq:mtrain}
\end{split}
\end{equation}
where $j$ is the index for the frequency bands. $\alpha$ is a uniformly distributed vector within $0$ to $K/T$. To calculate the phase shift $\textbf{S}$, motion features are embedded by a fully connected layer in each channel:
\begin{equation}
\begin{split}
  (s_x, s_y) = \bs{FC}(\textbf{y}_i),
  \textbf{S}_i = \bs{atant2}(s_y, s_x).
  \label{eq:mtrain}
\end{split}
\end{equation}
After acquiring the learned parameters $\textbf{A},\textbf{F},\textbf{B}$, and $\textbf{S}$, the corresponding latent space features are constructed using the following function:
\begin{equation}
\begin{split}
  \hat{\textbf{y}_{p}} = \textbf{A}\cdot \sin(2\pi \cdot (\textbf{F} \cdot T - \textbf{S})) + \textbf{B},
  \label{eq:mtrain}
\end{split}
\end{equation}
the reconstructed $\hat{\textbf{y}_{p}}$ has the same dimension as the input feature $\textbf{y}$.

\noindent \textbf{Individual non-periodicity.} As aforementioned, the periodic model is efficient in extracting periodic features from motion data, while the characteristics of human motions related to diverse audio input, and human identities are still neglected. Here we advocate to model these intrinsic characteristics by individual non-periodicity modeling. Toward this issue, we introduce an individual branch to enhance the disentanglement of non-periodic features in gestures, as depicted in \figref{fig:pipeline}. This branch incorporates an encoder, denoted as $\mathcal{E}_{np}$, which consists of multiple convolutional and normalization layers. The motion features $\textbf{y}$ are inputted into the encoder $\mathcal{E}_{np}$, resulting in the generation of latent non-periodic features $\hat{\textbf{y}}_{np}$. It is noteworthy that these extracted non-periodic features $\hat{\textbf{y}}_{np}$ maintain the same shape as the period embedding $\hat{\textbf{y}}_{p}$ obtained previously. This consistency in dimensionality facilitates subsequent fusion operations. To ensure the effectiveness of the extracted features,  the periodic features $\hat{\textbf{y}}_{p}$ are combined with non-periodic features $\hat{\textbf{y}}_{np}$ and then inputted into a decoder $\mathcal{D}_{bh}$, which is composed of a 1D convolutional layer, to reconstruct the target motion sequences,
\begin{equation}
\begin{split}
  \hat{\mc{C}}^{bh} = \mc{D}_{bh}(\hat{\textbf{y}_{p}} + \hat{\textbf{y}_{np}}),
  \label{eq:mtrain}
\end{split}
\end{equation}
$\hat{\mc{C}}^{bh}$ denotes the reconstructed gestures. In the periodic model, the loss function utilizes reconstruction loss to learn the distribution of motion data in space. To further capture the temporal correlations, a velocity loss is added between the input and reconstructed gestures to assess the performance of the model,
\begin{equation}
\begin{split}
  \mathcal{L}_{rec}^{bh} = &\underbrace{||\mc{C}^{bh} - \hat{\mc{C}}^{bh}||_1}_{\text{gesture motions}} + \lambda_u\underbrace{||\frac{\Delta (\mc{C}^{bh} - \hat{\mc{C}}^{bh})}{\Delta t}||_1}_{\text{gesture speed}},
\end{split}
\end{equation}
where $\lambda_u$ denotes the weight of the velocity loss of the body and hand gestures.

\subsection{Face Animation Generator}\label{sec:face}

In the research on the interaction between speech and facial expressions, our objective is to develop a face animation generator that can precisely synchronize with both the emotional content and spoken content of a given audio segment $\mc{C}_{1:T}^v$. The goal of this generator is to synthesize realistic facial animations $\mc{C}_{1:T}^m$ where there exists a strong one-to-one correspondence between lip movements and the verbal content within the audio, while non-lip facial movements are more closely tied to the emotional information conveyed in the audio. Unlike previous methods such as \cite{habibie2021learning, yi2023generating}, which primarily focused on extracting content-related features from audio and overlooked its emotional components, our approach utilizes pre-trained ASR and emotion classification models from Wav2vec 2.0 \cite{baevski2020wav2vec} to extract content and emotion features from speech. During training, the two pre-trained models $\mathcal{E}_{con}$ and $\mathcal{E}_{emo}$ based on extensive public audio datasets remain fixed, and the extracted content and emotion features are concatenated and fed into a face decoder for predicting corresponding face blendshapes $\hat{\mc{C}}_{1:T}^m$.

Considering that each person has their own habits, the speaker ID $\mc{C}^{id}$ and emotion label $\mc{C}^{emo}$ are encoded separately. To ensure consistency in the feature representation, both the ID and emotion label are encoded into a format of $\mathbb{R}^{8\times T}$. Furthermore, the transcript is also encoded, and these multimodal features are integrated with the audio. To make the feature information of different types have better interaction, instead of employing simple concatenation, two fully connected layers are utilized to process the features more deeply before inputting them into the decoder to generate the matching face animations. Due to the strong temporal dependencies in facial animations, the face decoder is designed with a bidirectional LSTM and three Temporal Convolutional Network (TCN) layers, followed by a fully connected layer to output 52 blendshape coefficients. This framework is trained with a combination of MSE and velocity losses to optimize the accuracy and smoothness of the generated animations.
\begin{equation}
  \hat{\mc{C}}_{1:T}^m = \mc{D}^m(\textbf{h}_{1:T}^v | \textbf{h}^{text}, \textbf{h}^{emo}, \textbf{h}^{id}),
  \label{eq:d_h}
\end{equation}
\begin{equation}
  \mc{L}_m = \omega_{mse}||\mc{C}_{1:T}^m - \hat{\mc{C}}_{1:T}^m||_2 + \omega_{vel}||\mc{C}_{1:T}^{m'} - \hat{\mc{C}}_{1:T}^{m'}||_1,
  \label{eq:l_m}
\end{equation}
\noindent where $\textbf{h}^{text}$, $\textbf{h}^{id}$, and $\textbf{h}^{emo}$ denote the features of the transcript, speaker ID, and emotion label respectively. $\textbf{h}^v$ is the concatenated content and emotion features of the audio. $\omega_{mse}$ and $\omega_{vel}$ denote the weights of the MSE and velocity losses respectively.

\subsection{Hierarchical Attribute Guidance}\label{sec:hie}

As indicated by the work \cite{liu2022beat}, researchers found that there exists an inter-connection between a character's gesture and his facial animations. Given the practical challenge of acquiring facial animations, a face animation generator is developed in \secref{sec:face} to extract facial features $\textbf{h}^m$ corresponding to the current audio. To intensify the influence of individual characteristics, the features of speaker ID $\textbf{h}^{id}$, emotion labels $\textbf{h}^{emo}$, and transcript $\textbf{h}^{text}$, all extracted in \secref{sec:face}, are utilized and fused with the facial features $\textbf{h}^m$.
In previous approaches, the audio features driving pose generation were extracted using MFCC or a content-based audio encoder, which couldn't accurately capture the emotional information related to the gestures in the audio. Therefore, the extracted audio embeddings $\textbf{h}^{v}$ are integrated with the features from other modalities and then fed into the established feature fusion network. To ensure the fused features $\textbf{h}_t^{fus}$ account for context, the feature fusion network is designed with one LSTM layer and two fully connected layers.
\begin{equation}
\begin{split}
  \textbf{h}_t^{fus} = \mc{E}_{fus}(\textbf{h}_t^{v}, \textbf{h}_t^{m}, \textbf{h}_t^{text}, \textbf{h}^{emo}, \textbf{h}^{id}|\textbf{h}_{t-1}^{fus}),~t = 1, 2, \dots, T
  \label{eq:fus}
\end{split}
\end{equation}
where $\mc{E}_{fus}$ denotes the feature fusion network, $\textbf{h}_{t-1}^{fus}$ denotes the multimodal representation of features fused at the $t-1$ time step.

\noindent \textbf{Gesture Generator.} In the motion inference phase, drawing inspiration from \cite{liu2022beat,ng2021body2hands}, which highlights the Inter-correlation of body movements on hand gestures, a cascaded structure is designed to synthesize the body $\mc{C}^{body}$ and hand gestures $\mc{C}^{hand}$. The body and hand generator utilizes a TCN to encode the fused multi-modal features, mapping them to the corresponding poses through a fully connected layer. Since body gestures are simpler compared to hand movements, two layers of TCN and one fully connected layer are employed to generate the body gestures $\hat{\mc{C}^{body}}$ that align with the speech.

After obtaining the body poses, the multimodal features are combined with them to predict the hand gestures. To enhance the model's ability to generate complex hand motions, the hand generator is constructed with four layers of TCN and one fully connected layer. The fused features are then input into the hand decoder to predict hand gestures that align with the audio and body. Finally, the body and hand gestures are combined to obtain the complete upper body motions.
\begin{equation}
\begin{split}
  \hat{\mc{C}_{1:T}^{hand}} = \mc{D}_{hand}(\textbf{h}_{1:T}^{fus} | \hat{\mc{C}_{1:T}^{body}}).
  \label{eq:mtrain}
\end{split}
\end{equation}
In the training process, the reconstruction and velocity losses $\mathcal{L}_{rec}^{ges}$ are measured between the synthesized results and the original inputs. The loss $\mathcal{L}_{rec}^{ges}$ is similar to the $\mathcal{L}_{rec}^{bh}$.

\noindent \textbf{Gesture Enhancement.} During the co-speech gesture training phase, the Weight-Blended Mixture-of-Experts framework proposed in \cite{starke2020local} is adopted as the period-element enhancer for the PD module in \secref{sec:PD}. Both gating $\mc{E}_w$ and expert $\mc{D}_w$ networks are comprised of three fully connected layers. In contrast to the work \cite{starke2022deepphase}, which predicted the corresponding periodic parameters $\textbf{A}, \textbf{F}, \textbf{B}, \textbf{S}$ frame by frame, multiple-frame periodic signals are simultaneously predicted by combining the extracted multi-frame multimodal features. Due to the temporal dependencies in the phase, LSTM is adopted to construct the periodic decoder. The obtained symbols are then input into the gates $\mc{E}_w$ to predict the weights of different experts $\mc{D}_w$. Then, the fused multi-modal features $\textbf{h}_t^{fus}$ are fed into the expert layer to synthesize the corresponding period element of the gestures. Finally, the generated motions $\hat{\mc{C}_{1:T}^{body}}$ and $\hat{\mc{C}_{1:T}^{hand}}$ are combined with the periodic elements to synthesis the final results. The overall loss $\mc{L}$ is presented as:
\begin{equation}
\begin{split}
  \mc{L} = \mc{L}_{rec}^{ges} + \sum_{\mc{Z}}||\mc{Z} - \hat{\mc{Z}}||^2_2, \mc{Z} \in \{\textbf{A}, \textbf{F}, \textbf{B}, \textbf{S}\},
  \label{eq:mtrain}
\end{split}
\end{equation}
where $\mc{Z}$ and $\hat{\mc{Z}}$ denote the pseudo and predicted periodic parameters. During the inference phase, the model first generates facial animation $\hat{\mc{C}^{m}}$ corresponding to the audio $\mc{C}^{v}$ based on multimodal data. The extracted facial information is then fused with the multi-modal features and separately input into the gesture generator and periodic enhancement module to synthesize the non-periodic and periodic components of the motions. Finally, these components are combined to obtain the body and hand gestures.

\section{Experiments}\label{sec:exp} 

\begin{figure*}[t]
\begin{center}
   \includegraphics[width=0.8\linewidth]{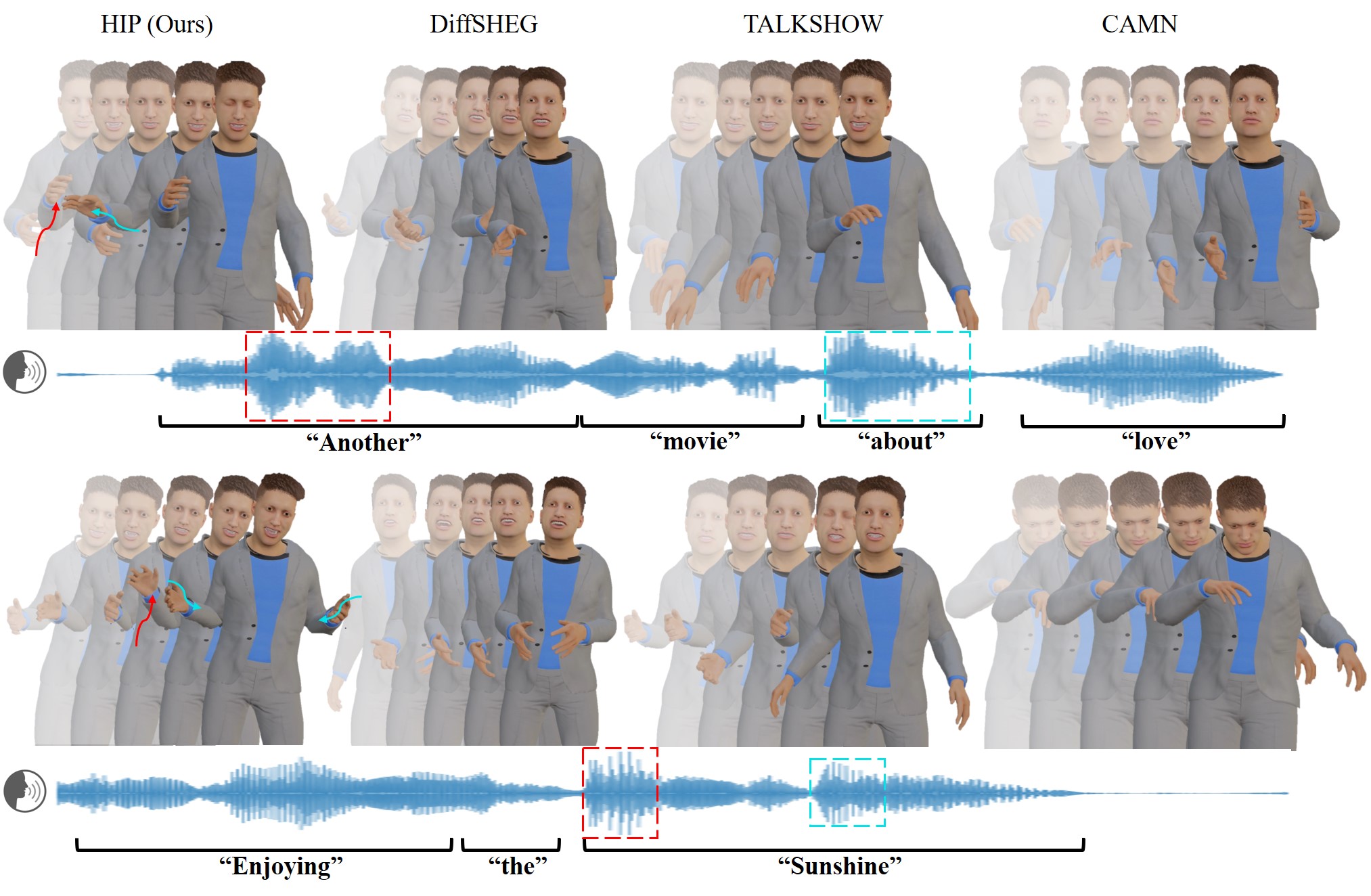}
\end{center}
   \caption{Comparison of results generated by different methods. The curve represents the hand movement trajectory, and the boxes highlight audio segments with noticeable rhythmic fluctuations. Compared to other methods, our model generates movements that better align with both the semantics and rhythm of the speech. When saying ``another" with a clear pitch fluctuation, our model flips the wrist upwards (\textcolor{red}{red}). When saying ``sunshine", the hand first moves upwards (\textcolor{red}{red}) and then comes together (\textcolor{cyan!50}{blue}), with the movement speed closely matching the rhythm changes in the speech.}
\label{fig:comges}
\end{figure*}

\renewcommand{\arraystretch}{1.2}
\setlength{\tabcolsep}{14pt}

\begin{table*}[!h]
  \centering
  \caption{Quantitative comparisons on the BEAT dataset between our method and other works. The best values are bolded while the second-place performances are underlined. $\downarrow$:The lower the better. $\uparrow$: The larger the better. $\textbf{FGD}$, $\textbf{SRGR}$, $\textbf{Diversity}$ and $\textbf{BeatAlign}$ are computed using the officially evaluation code from CAMN~\cite{liu2022beat}.}
  \label{tab:compare}
  \begin{tabular}{l|c|cccc}\hline
    \toprule
    Method & Input Modalities & $\textbf{FGD}$ $\downarrow$ & $\textbf{SRGR}$ $\uparrow$ & $\textbf{Diversity}$ $\uparrow$ & \textbf{BeatAlign} ({BA}) $\uparrow$ \\
    \hline
    S2G~\cite{ginosar2019learning} & audio & 232.6 & 0.133 & 10.33 & 0.725 \\
    Trimodal~\cite{yoon2020speech} & audio, text &  176.2 & 0.196 & 12.17 & 0.766 \\
    Habibie~\etal~\cite{habibie2021learning} & audio & 183.2 & 0.208 & \underline{13.05} & 0.730 \\
    A2G~\cite{li2021audio2gestures} & audio & 125.8 & 0.192 & 10.52 & 0.767 \\
    CAMN~\cite{liu2022beat} & audio, text, facial& 91.3 & 0.259 & 12.86 & 0.779 \\
    TALKSHOW~\cite{yi2023generating} & audio & 106.4 & 0.271 & 11.92 & 0.774 \\
    DiffSHEG~\cite{chen2024diffsheg} & audio & \underline{85.2} & \underline{0.275} & 11.35 & \textbf{0.791} \\
    \bottomrule
    HIP (Ours) & audio, text & \textbf{70.9} & \textbf{0.283} & \textbf{13.51} & \underline{0.787} \\
    \hline
  \end{tabular}
\end{table*}

\begin{table}[!h]
  \centering
  \setlength{\tabcolsep}{8pt}
  \caption{Quantitative results on the BEATv2 dataset~\cite{liu2023emage}. \textbf{FGD}, \textbf{Diversity}, and \textbf{BeatAlign} are computed using the official evaluation code released with EMAGE~\cite{liu2023emage}.}
  \label{tab:compare2}
  \begin{tabular}{l|ccc}
    \hline
    \toprule
    Method & \textbf{FGD} $\downarrow$ & \textbf{Diversity} $\uparrow$ & \textbf{BeatAlign} ({BA}) $\uparrow$ \\
    \hline
    ProbTalk~\cite{liu2024towards} & 5.686 & 11.84 & 7.490 \\
    EMAGE~\cite{liu2023emage} & 5.512 & \underline{13.06} & 7.724 \\
    MambaTalk~\cite{xu2024mambatalk} & \underline{5.366} & 13.05 & \underline{7.812} \\
    \bottomrule
    HIP (Ours) & \textbf{5.293} & \textbf{13.11} & \textbf{7.948} \\
    \hline
  \end{tabular}
\end{table}

\noindent \textbf{Dataset:} We conduct our experiments on the BEAT~\cite{liu2022beat} and BEATv2~\cite{liu2023emage} datasets, following the settings adopted in previous works~\cite{liu2022beat,liu2023emage}: four English-speaking speakers from BEAT and one English-speaking speaker from BEATv2, both providing synchronized audio, text, speaker IDs, facial animations, and full-body motion capture data.

\noindent \textbf{Implementation Details:} Following official configurations, BEAT~\cite{liu2022beat} motion data is downsampled to 15 fps, while BEATv2~\cite{liu2023emage} is at 30 fps. Our overall framework is trained on a single consumer-level NVIDIA RTX 3090 GPU, 16 cores CPUs, and 32GB memory. During training, the motion length in the samples is 34, with a batch size of 256. The number of channels in the period model is set to 10. We utilize the Adam optimizer with a learning rate of 5.0e-4, training the model for 200 epochs.

\begin{figure*}[t]
\begin{center}
   \includegraphics[width=0.8\linewidth]{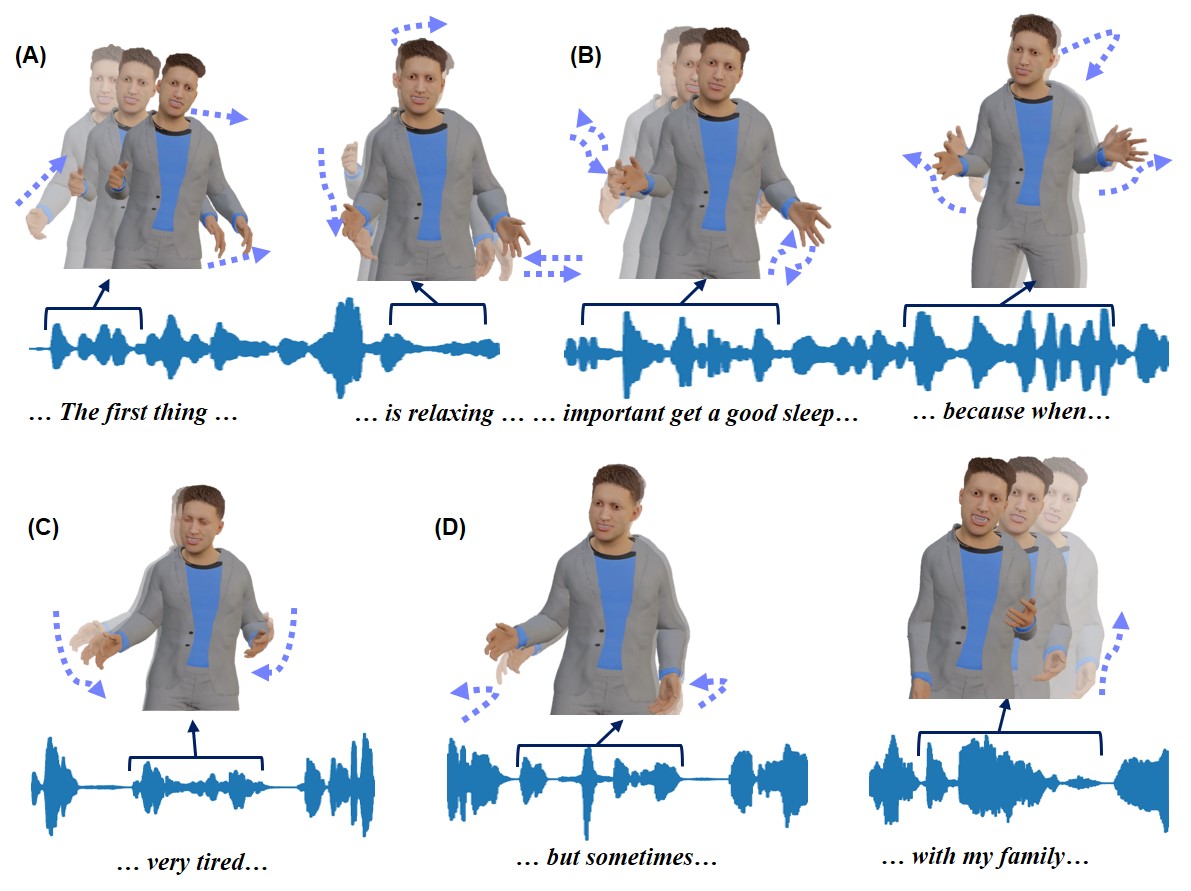}
\end{center}
   \caption{The sample results of co-speech gesture generated from ours. It includes motion trajectories, speech, and text. (A) When saying ``The first thing," the character makes a preparatory gesture. There is a swinging motion when saying ``relaxing." (B) There are rhythmic hand gestures and metaphorical gestures when the character speaks quickly and says ``when." (C) The character makes a lowering hand gesture when saying ``tired." (D) There is a metaphorical gesture when saying ``sometimes" and ``my family."}
\label{fig:diverse}
\end{figure*}

\noindent \textbf{Evaluation Metrics:} To evaluate the performance of our method in generating facial animations, we calculate the mean squared error between predicted and ground truth values from two aspects: lip average distance (LAD) and facial average distance (FAD). To assess the quality of the gestures generated by ours, we follow the work~\cite{liu2022beat,liu2023emage} to evaluate the results in several aspects, which includes assessing the quality of the generated gestures (FGD), the correlation between motions and semantics (SRGR), and the beat alignment between the audio and the generated results. Specifically, FGD measures the quality by computing the distance between the features of generated and ground-truth movements. The features of gestures are extracted using a pre-trained auto-encoder model that is trained on the gesture data. SRGR evaluates the correlation between the generated motions and semantics by using the Probability of Correct Keypoint (PCK). Specifically, PCK measures joint accuracy by comparing the number of correctly recalled joints against a specific threshold $\sigma$.
\begin{equation}
\begin{split}
  \mc{S}_{\bs{SRGR}} = \lambda\sum{\frac{1}{T\times J}}\sum_{t=1}^{T}\sum_{j=1}^{J}\mathbbm{1}[||g_t^j-\hat{g}_t^j]||_2 < \sigma],
  \label{eq:srgr}
\end{split}
\end{equation}
where $\hat{g}$ denotes the predicted joints and $g$ denotes the ground truth. $\mathbbm{1}(\cdot)$ is the indicator function and $J$ is the number of joints. BeatAlign assesses the Chamfer Distance between audio and gesture beats to evaluate the similarity between the rhythm of the gesture and audio.
\begin{equation}
  \textbf{BA}= \frac{1}{|\textbf{B}_v|}\sum\limits_{\textbf{v}_i \in \textbf{B}_v}\exp(- \frac{\min_{\textbf{g}_i \in \textbf{B}_g}||\textbf{g}_i-\textbf{v}_i||^2}{2\tau^2}),
  \label{eq:ba}
\end{equation}
where $\textbf{B}_v$ and $\textbf{B}_g$ denote the beat of speech and gestures. $\tau$ denotes the normalized parameter. 

\subsection{Quantitative Evaluation}

In our experiments, we compare our method with state-of-the-art approaches, including S2G~\cite{ginosar2019learning}, Trimodal~\cite{yoon2020speech}, Habibie~\cite{habibie2021learning}, A2G~\cite{li2021audio2gestures}, CAMN~\cite{liu2022beat}, TALKSHOW~\cite{yi2023generating}, and DiffSHEG~\cite{chen2024diffsheg}. As shown in \tabref{tab:compare}, our method achieves the best performance in FGD and SRGR on the BEAT dataset, validating that explicitly disentangling periodic and non-periodic components significantly enhances the realism and semantic alignment of the generated gestures. Our method obtains slightly lower BeatAlign than DiffSHEG, which tends to generate more short-term transient movements detected as additional beats, even when temporal fluctuations are present. In contrast, our model focuses on producing physically plausible and semantically coherent gestures, resulting in fewer spurious high-frequency spikes and thus slightly lower BeatAlign but better overall quality. To further verify the robustness of our approach, we also evaluate it on BEATv2~\cite{liu2023emage}, and compare against ProbTalk~\cite{liu2024towards}, EMAGE~\cite{liu2023emage}, and MambaTalk~\cite{xu2024mambatalk}. As shown in \tabref{tab:compare2}, the results show that our method consistently achieves the best scores in FGD, Diversity, and BeatAlign, demonstrating that it remains effective and robust when evaluated on BEATv2.

\noindent \textbf{Gesture Visualization}: To better compare our results with other methods, we visualize the results generated from CAMN \cite{liu2022beat}, TALKSHOW \cite{yi2023generating}, DiffSHEG \cite{chen2024diffsheg} and ours in \figref{fig:comges}. As shown in \figref{fig:comges}, although CAMN generates continuous motions, the speed is slow and the diversity is poor. TALKSHOW, based on quantized encoding, can generate more diverse motions, but the generated motions tend to be simplistic and exhibit jitter. Although DiffSHEG shows improvements in diversity and motion complexity, it still produces some jittering movements. Compared to the methods mentioned above, our approach generates continuous, diverse, and realistic motions that better align with the content and rhythm of the speech.

\noindent \textbf{Gesture Consistency}: To verify the alignment of our generated poses with the semantic content and rhythm aspects of the audio, we visualize the generated results alongside their corresponding audio and text, as shown in \figref{fig:diverse}. The character makes metaphorical gestures when saying ``relaxing," ``when," ``tired," and ``my" in \figref{fig:diverse} (A), (B), (C), and (D). A starting gesture appears in \figref{fig:diverse} (A) when saying ``the first." This result demonstrates that the results generated from ours align with the content of the audio. When speaking quickly, real human movements often exhibit swinging or no movement. In \figref{fig:diverse} (B), when speaking rapidly, the character makes a simple swinging motion to match the current speech rhythm. This result further validates that the gestures generated from ours well match the rhythm of the speech.

\begin{table}[!t]
  \centering
  \caption{Quantitative comparisons on the task of the face animations generation of the BEAT dataset.}
  \label{tab:face}
  \begin{tabular}{l|cc}
    \toprule
    Method & $\textbf{LAD}$ $\downarrow$ & $\textbf{FAD}$ $\downarrow$ \\
    \midrule
    Habibie \etal~\cite{habibie2021learning} & 0.072 & 0.714\\
    TALKSHOW \cite{yi2023generating} & 0.046 & 0.530\\
    DiffSHEG \cite{chen2024diffsheg} & 0.043 & 0.412\\
    \hline
    Ours & \textbf{0.042} & \textbf{0.361}\\
    \bottomrule
  \end{tabular}
\end{table}

\begin{figure}[!t]
\begin{center}
   \includegraphics[width=1.0\linewidth]{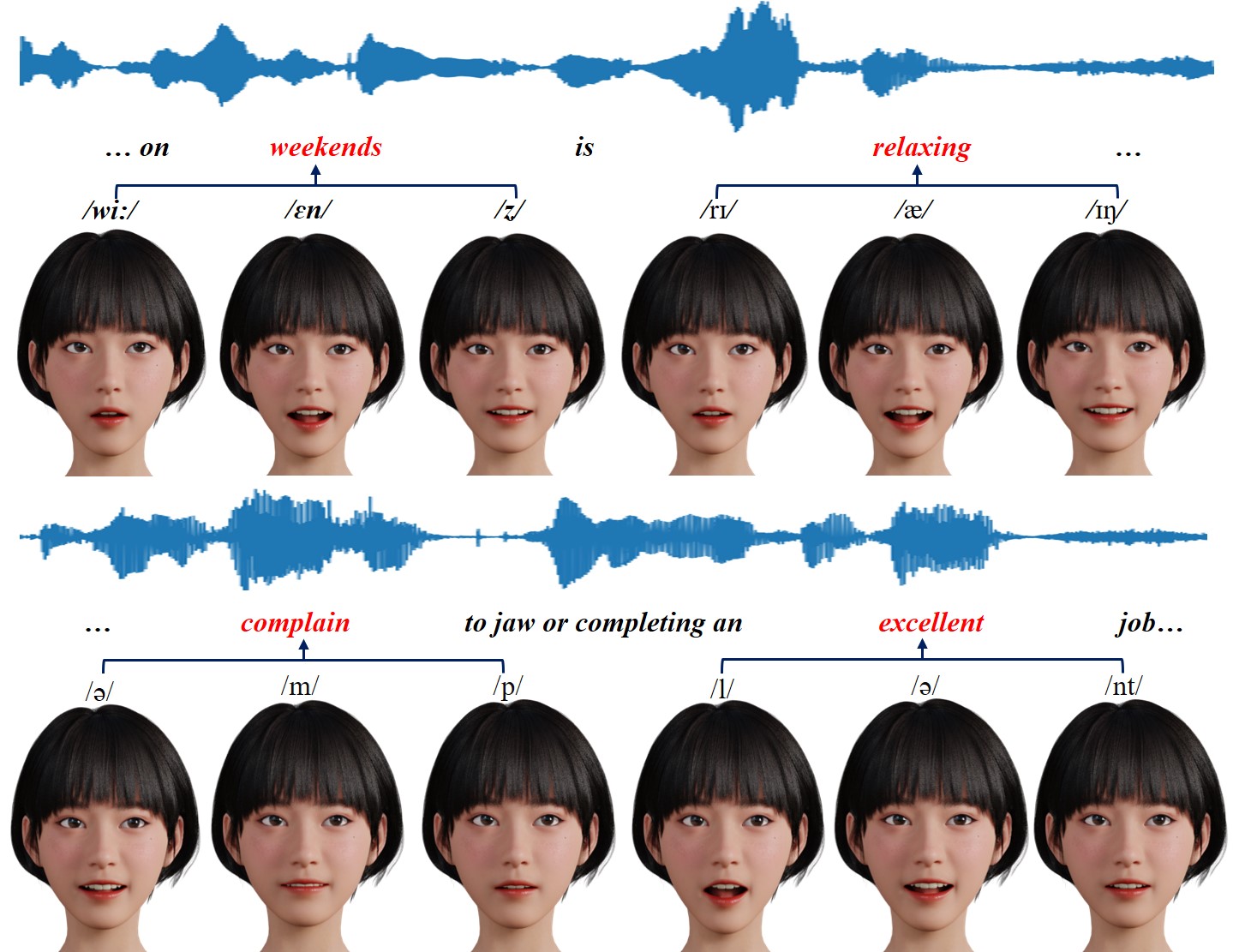}
\end{center}
   \caption{Examples of generated face animations from ours. The results generated from ours match the audio with continuous and accurate lip motions.}
\label{fig:audioface}
\end{figure}
 
\noindent \textbf{Face Animations:} The performance of face animation generation is compared in \tabref{tab:face}. Our model outperforms Habibie \cite{habibie2021learning}, TALKSHOW \cite{yi2023generating}, and DiffSHEG \cite{chen2024diffsheg} in two evaluation metrics. The results are further visualized in \figref{fig:audioface}, showcasing animations generated for different phonemes. Our generator produces facial animations that closely resemble real facial expressions. For example, when pronouncing /w/, the lips move towards the center; for /z/, the lips spread out to the sides; and during the sounds /m/ and /p/, the lips close slightly. These results demonstrate the capability of our model to accurately capture nuanced lip movements corresponding to different phonemes, enhancing the realism of the generated facial animations.

\begin{figure}[t]
\begin{center}
   \includegraphics[width=1.0\linewidth]{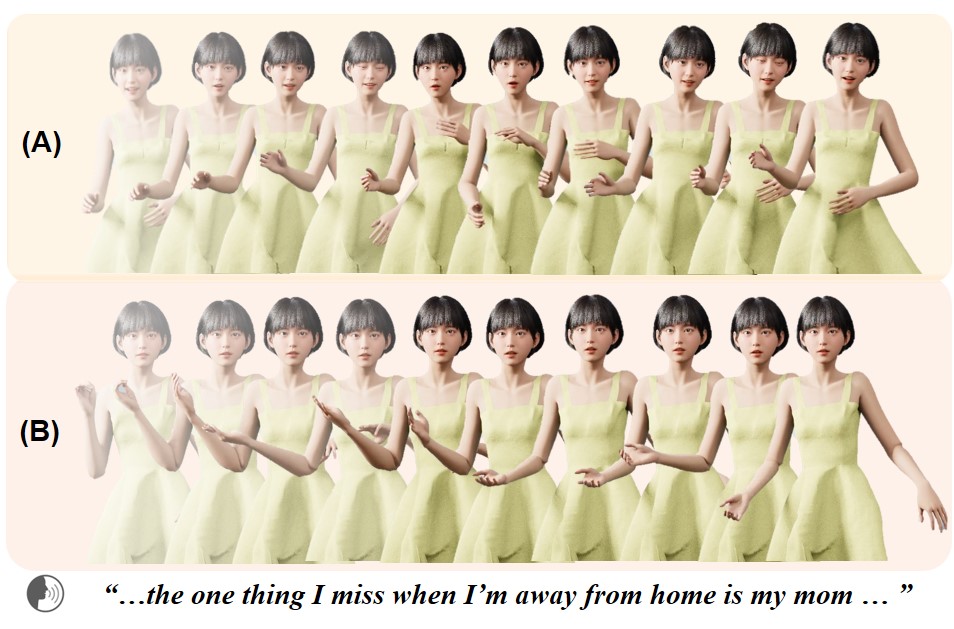}
\end{center}
   \caption{Co-speech gestures from different speakers with the same text. The results of (A) and (B) are two different characters from the test set. When they say the same content, the generated poses exhibit different trajectories.}
\label{fig:difspeaker}
\end{figure}

\noindent \textbf{Person Diversity}: We visualize the facial and body animations generated by two different characters saying the same text, as shown in \figref{fig:difspeaker}. In the facial animation, (A) exhibits larger mouth movements, while (B) shows smaller mouth movements. This phenomenon indicates that our facial animation module effectively learns the facial styles of different characters when speaking. In terms of body gestures, (A) has more head and spinal movements than (B), and the hand movements also follow different trajectories. This validates that after encoding different person IDs, the model further combines the distinguishable information for each character, such as facial features, and maps it to the corresponding poses, effectively learning the pose styles of different characters.

\begin{figure}[t]
\begin{center}
   \includegraphics[width=1.0\linewidth]{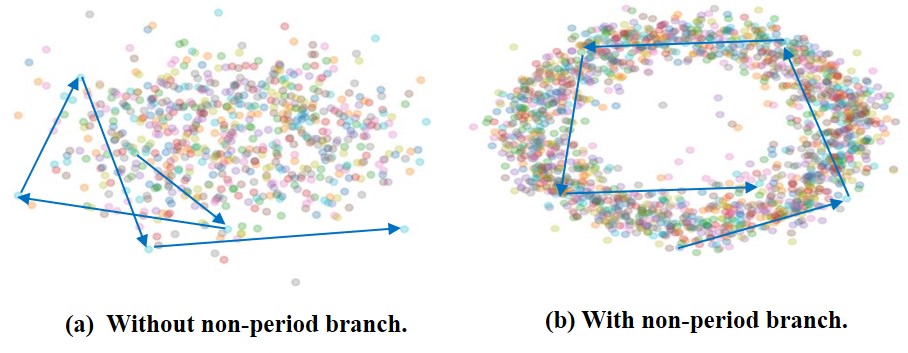}
\end{center}
   \caption{The effectiveness of periodic feature extraction after using our method. The phase spaces are visualized by 2D PCA projection. The blue line indicates adjacent points and directions. (a) The results are without the non-period branch. (b) The results after implementing the non-period branch. }
\label{fig:compae}
\end{figure}

\noindent \textbf{Parameter Visualization}: The periodic parameters extracted from the gestures by DeepPhase \cite{starke2022deepphase} and our method are visualized in \figref{fig:compae}. As can be seen from the \figref{fig:compae} (a), before adding any extra channels, the extracted parameters present an irregular distribution. This proves that directly extracting the periodic features from the gestures has a bad effect. However, in the \figref{fig:compae} (b), we find that after adding the non-period branch, the extracted results show a more regular distribution. When there are no non-periodic channels, both the periodic and non-periodic information in the data are extracted and reconstructed through separate periodic channels. Due to the influence of non-periodic features, the model cannot recognize the periodic components in the data well. However, by using two branches, one using periodic functions to drive the model to extract periodic features, and the other using convolution to extract non-periodic embeddings, the outputs are then added together and the validity of the feature extraction is verified through reconstruction. Therefore, the model can successfully disentangle the periodic and non-periodic features of the data, effectively improving the model's capabilities.

\begin{figure}[!t]
\begin{center}
   \includegraphics[width=1.0\linewidth]{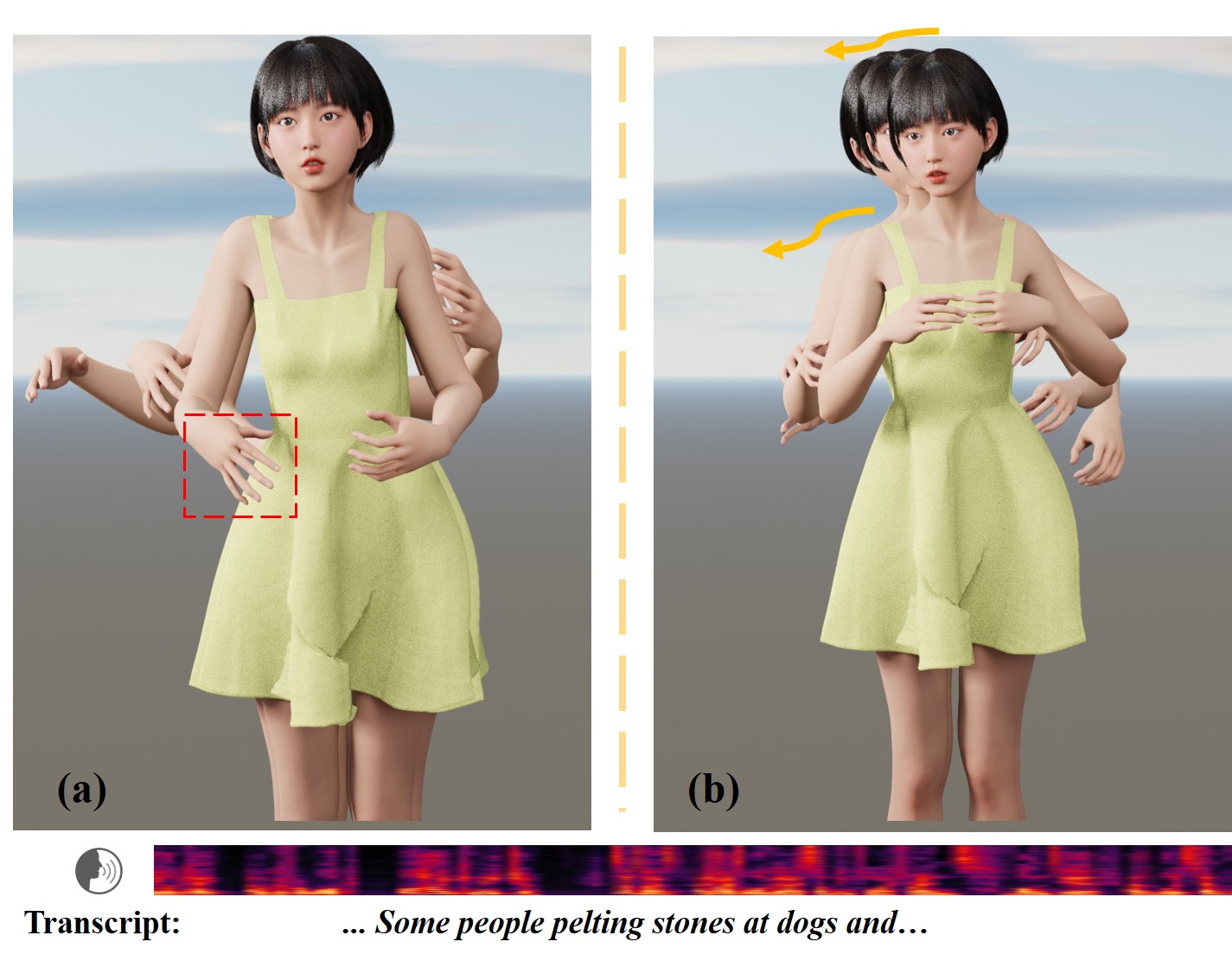}
\end{center}
   \caption{Examples of different versions of CAMN generating poses with the same input. (a) The pose generated without the periodicity disentanglement module; (b) The motion generated after adding the periodicity disentanglement module.}
\label{fig:combeat}
\end{figure}

\begin{table}[!t]
  \centering
  \caption{{\bf Cross validation about CAMN \cite{liu2022beat} on BEAT dataset}. The $\mathcal{P}$ denotes the periodicity disentanglement module.}
  \label{tab:additional}
  \begin{tabular}{l|cc}
    \toprule
    Method & \textbf{FGD} $\downarrow$ & $\Delta_{imp}$ \\
    \midrule
    CAMN & $91.3$ & \\
    CAMN w. $\mathcal{P}$ & $84.3$ & $(-7.0)$\\
    \bottomrule
  \end{tabular}
\end{table}

\noindent \textbf{Module Generalization}: To validate the effectiveness of the periodicity disentanglement module in improving the quality of the generated gestures, we attempt to strengthen the CAMN \cite{liu2022beat} by incorporating it. The improved results are shown in the \tabref{tab:additional}. After adding the periodicity disentanglement module, the quality of the gestures generated from the model shows obvious improvements. The reason for this phenomenon is that the goal of the origin model is to converge the loss, so it tends to generate the principal components of the gestures during the training phase. Therefore, directly constructing the mapping relationship between audio and gestures often results in generating rigid and unnatural motions. However, the periodicity disentanglement module drives the model to further learn the periodic patterns in the actions, making the generated results more realistic. To observe the improvement in the naturalness of motions with the proposed module, the generated results under different versions are visualized in \figref{fig:combeat}. As can be seen from the results, the gestures of the body generated from CAMN \cite{liu2022beat} have little physical change, and the hand in the red box appears unnatural. After adding the periodicity disentanglement module, the spine and head of the character exhibit small movements like a human's as the voice changes. 

\begin{table}[!t]
  \centering
  \caption{\textbf{Ablation study about our method on BEAT dataset}. The $\mathcal{F}$ represents facial information in the fusion features. $\mathcal{NP}$ and $\mathcal{N}$ refer to the non-periodic branch and the number of channels in the periodicity disentanglement module $\mathcal{P}$, respectively.}
  \label{tab:ablation}
  \begin{tabular}{ccc|cc}
    \toprule
    \multicolumn{3}{c|}{Model Variations} & \multirow{2}{*}{\textbf{FGD} $ \downarrow$}  & \multirow{2}{*}{$ \Delta_{imp}$}\\
    \cline{1-3}
    $\mathcal{F}$ & $\mathcal{P}$ & $\mathcal{N}$ & ~ & ~ \\
    \midrule
    $\times$ & $\times$ & $\times$ & $119.3$ & \\
    \checkmark & $\times$ & $\times$ & $97.2$ &$(-22.1)$ \\
    $\times$ & w. $\mathcal{NP}$ & 10& $89.4$ &$(-29.9)$ \\
    \checkmark & w/o. $\mathcal{NP}$ & 10 & $96.5$ & $(-22.8)$ \\
    \checkmark & w. $\mathcal{NP}$ & 2 & $79.7$ &  $(-39.6)$ \\
    \checkmark & w. $\mathcal{NP}$ & 5 & $72.5$ &  $(-46.8)$ \\
    \hline
    \checkmark & w. $\mathcal{NP}$ & 10 & $\textbf{70.9}$ & $(-48.4)$\\
    \bottomrule
  \end{tabular}
\end{table}

\subsection{Ablation Study}

To evaluate the effectiveness of our proposed method, we conduct an ablation study on our model. Specifically, we set up several model variants that differ from our approach. From the results in the \tabref{tab:ablation}, we find that in the absence of the face features $\mathcal{F}$, FGD is reduced by $18.5$ compared to our model. This indicates that face information is beneficial for motion generation, further validating the effectiveness of our proposed method. After adding the face features, our work further improves the similarity between generated and ground-truth actions. In the presence of the periodicity disentanglement module $\mathcal{P}$, FGD improves by $22.1$, proving that this module effectively enhances the quality and naturalness of the generated gestures. When the non-periodic branch $\mathcal{NP}$ in the periodicity disentanglement module is removed, the extracted periodic information does not significantly improve the quality of the generated motion. However, as the number of channels in the periodicity disentanglement module increases, the quality of the generated motion improves further. These results further validate the effectiveness of facial information and the extracted periodic information in improving the quality of the generated motion.

\begin{table}[!t]
  \centering
  \caption{{\bf User study about our method on BEAT dataset}. The table shows the percentage of user preferences for different methods and our method based on the two metrics: \textbf{Realism} and \textbf{Gesture-Speech Sync}. Higher values indicate better results in comparison to the given methods.}
  \label{tab:user}
  \begin{tabular}{c|cc}
    \toprule
    Method & \textbf{Realism} & \textbf{Gesture-Speech Sync}\\
    \midrule
    Habibie \etal~\cite{habibie2021learning} & 81.6\% & 75.4\% \\
    CAMN \cite{liu2022beat} & 75.6\% & 72.7\% \\
    TALKSHOW \cite{yi2023generating} & 66.3\% & 70.2\% \\
    DiffSHEG \cite{chen2024diffsheg} & 56.5\% & 54.8\% \\
    \hline
    HIP (Ours) & - & - \\
    \bottomrule
  \end{tabular}
\end{table}

\subsection{User Study}

To further validate the quality of the co-speech gestures generated from ours, we conduct a user study with recent representative methods, ~\ie, Habibie \cite{habibie2021learning}, CAMN \cite{liu2022beat}, TALKSHOW \cite{yi2023generating}, DiffSHEG \cite{chen2024diffsheg}. Specifically, 13 volunteers, including 6 men and 7 women, are invited, and a platform is designed for the participants to watch videos and select the results they consider better.

To ensure fairness in the experiment, the same audio and text are input for each method to generate the corresponding pose sequences. $20$ pairs of generated videos are randomly played, with one video generated by our method and the other randomly selected from the other works. Participants are unaware of which video is generated by us before evaluating them. As shown in \tabref{tab:user}, compared to Habibie and CAMN, over $70\%$ of the results generated by our model are considered superior to theirs in both metrics. According to their feedback, although the generated movements do not exhibit teleportation, the body movements are simple and stiff, with the body remaining upright and still, making them look unnatural, which affects their scores. Compared to TALKSHOW, our method generates body movements with higher complexity, and the body in our results is also more diverse. Although DiffSHEG has made some progress, certain instances of misalignment with speech and jitter affect its score. These results further verify that our proposed method can generate more realistic character movements.

\subsection{Limitation}
Our framework introduces a hierarchical architecture for generating full-body animations, producing facial and body motions that match speech. While effective, it has two main limitations. First, the extraction of periodic features plays an important role in generating body gestures, yet a few failure cases remain (\figref{fig:compae}~(b)). The choice of channel dimensionality in the disentanglement module also affects performance: increasing the number of channels can enhance periodic patterns but often leads to overfitting, thereby degrading the quality of the generated gestures. Second, the dependencies between different gesture units are currently modeled in an implicit manner, which limits the transparency of the generation process and hinders intuitive control over how different parts influence each other. We will explore more interpretable modeling strategies to explicitly represent these dependencies in future work.

\section{Conclusions}\label{sec:conclusion}

In this paper, we explore the implicit rules of co-speech human gesture movements and start a different insightful view to model this learning process compared to prevailing literature. To fulfill this generation process, we first propose a periodicity disentanglement module to model the regular periodic phase manifold as well as the non-periodic individual latent. We then build a face animation generator and construct hierarchical attribute guidance to implicitly model the inter-relationship of the human face, body, and hand gestures. Despite the significant experimental improvements and verifications, our proposed method models the learning relationship of multiple gesture units in an implicit manner, while the concrete explicit correlations still need further exploration.

\section*{Acknowledgment}
This work is partially supported by grants from the National Natural Science Foundation of China (No. 62132002 and No. 62202010), Guizhou Provincial Major Scientific and Technological Program (Qiankehe Zhongda [2025] No. 032), Beijing Nova Program (No. 20250484786), and the Fundamental Research Funds for the Central Universities.
\ifCLASSOPTIONcaptionsoff
  \newpage
\fi

\ifCLASSOPTIONcaptionsoff
  \newpage
\fi

\bibliographystyle{IEEEtran}
\bibliography{egbib}

\begin{IEEEbiography}[{\includegraphics[width=1in,height=1.25in,clip,keepaspectratio]{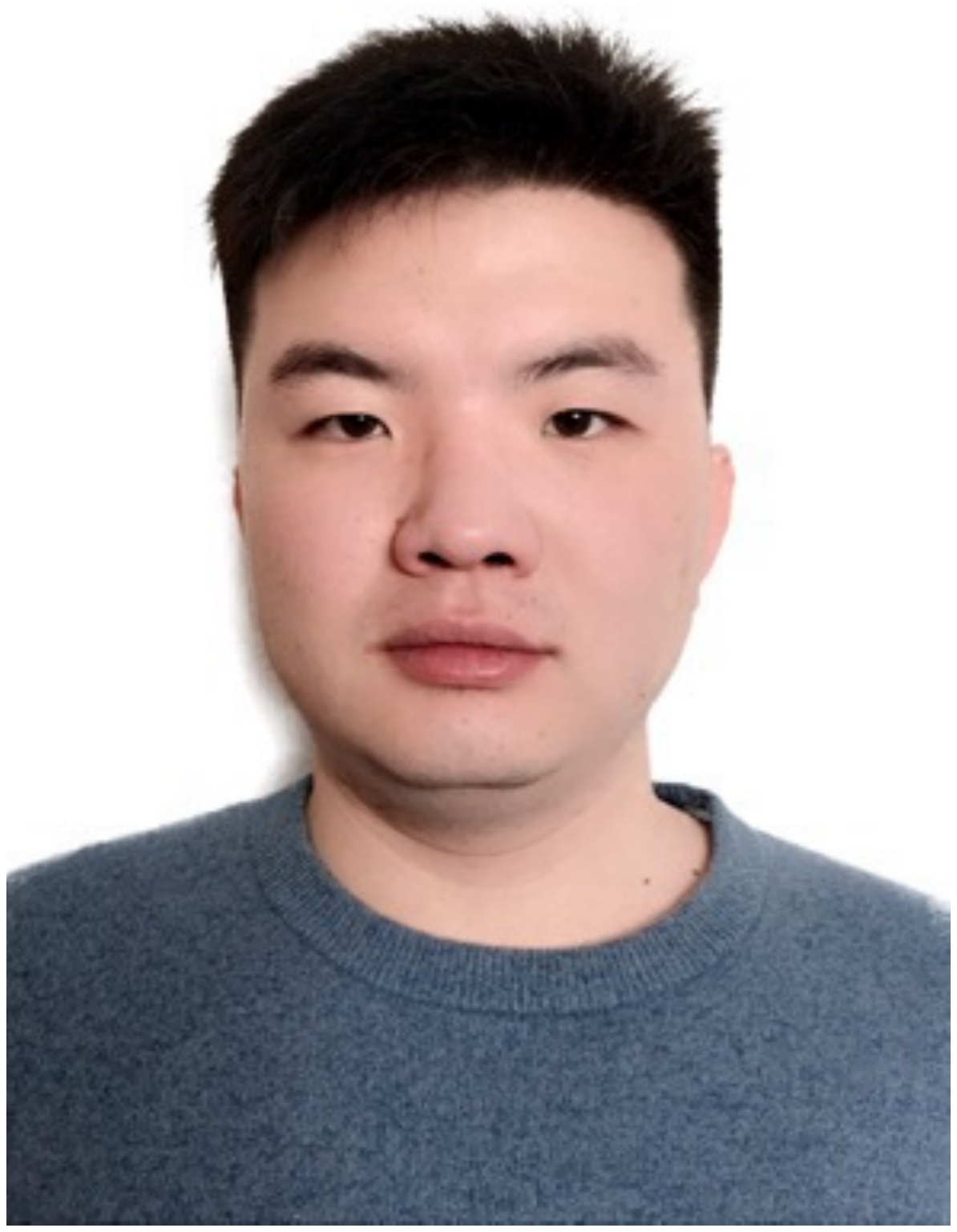}}]{Xin Guo} is currently pursuing the Ph.D. degree with the State Key Laboratory of Virtual Reality Technology and System, School of Computer Science and Engineering, Beihang University. His research interests include computer vision and cross-modal learning.
\end{IEEEbiography}

\begin{IEEEbiography}[{\includegraphics[width=1in,height=1.25in,clip,keepaspectratio]{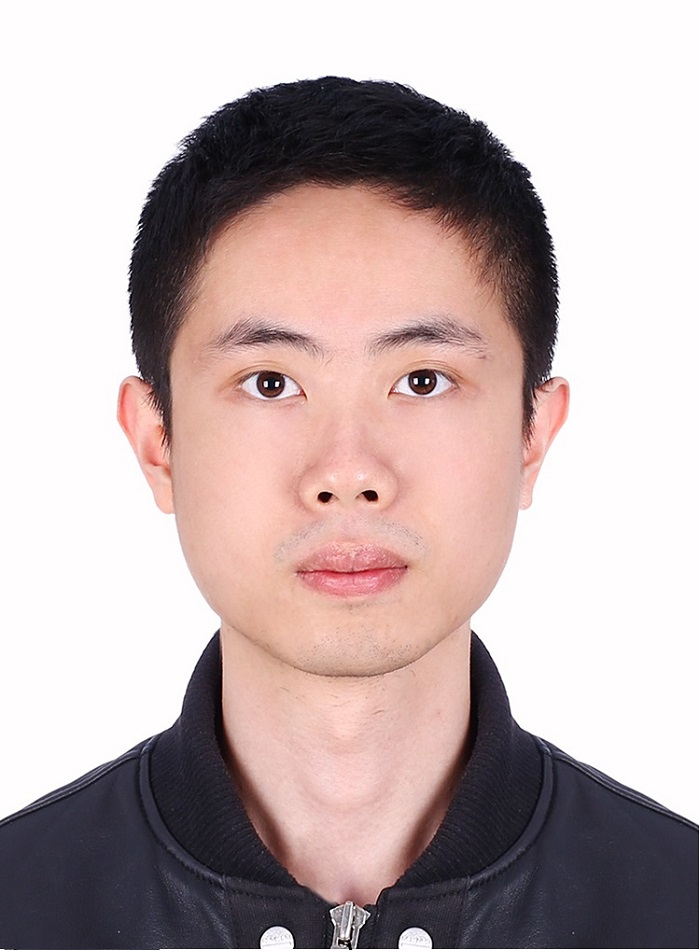}}]{Yifan Zhao} (Member, IEEE) is currently an Associated Professor with the School of Computer Science and Engineering, Beihang University, Beijing, China. He worked as a Boya Postdoc researcher with the School of Computer Science, Peking University. He received the B.E. degree from the Harbin Institute of Technology in Jul. 2016 and the Ph.D. degree from the School of Computer Science and Engineering, Beihang University, in Oct. 2021. His research interests include computer vision, VR/AR, and image/video understanding.
\end{IEEEbiography}

\begin{IEEEbiography}[{\includegraphics[width=1in,height=1.25in,clip,keepaspectratio]{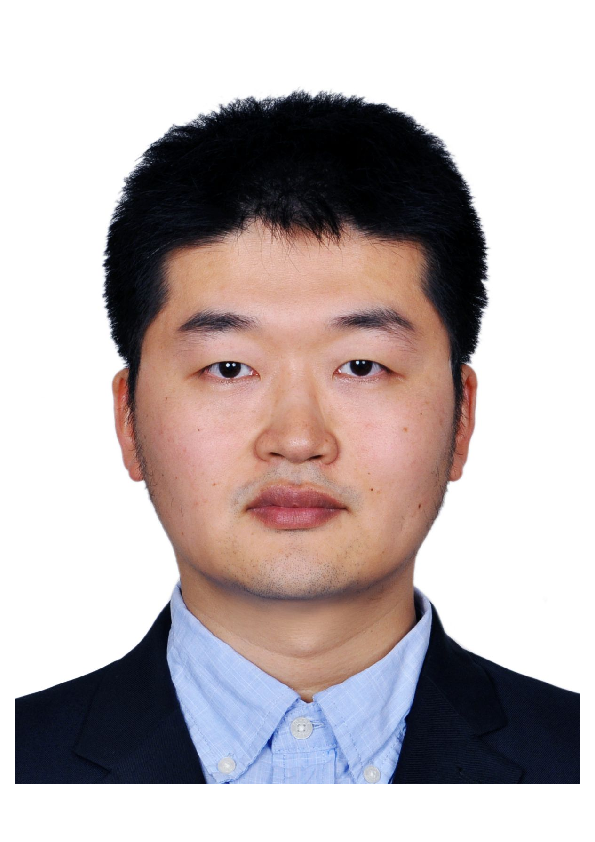}}]{Jia Li} (M'12-SM'15) received the B.E. degree from Tsinghua University in 2005 and the Ph.D. degree from the Institute of Computing Technology, Chinese Academy of Sciences, in 2011. He is currently a Full Professor with the School of Computer Science and Engineering, Beihang University, Beijing, China. He is the author or coauthor of over 100 technical articles in refereed journals and conferences such as TPAMI, IJCV, TIP, CVPR and ICCV. His research interests include computer vision and multimedia big data, especially the understanding and generation of visual contents. He is supported by the Research Funds for Excellent Young Researchers from the National Nature Science Foundation of China since 2019. He was also selected into the Beijing Nova Program (2017) and ever received the Second-grade Science Award of Chinese Institute of Electronics (2018), two Excellent Doctoral Thesis Awards from Chinese Academy of Sciences (2012) and the Beijing Municipal Education Commission (2012), and the First-Grade Science-Technology Progress Award from Ministry of Education, China (2010). He is an IET Fellow, and a Senior Member of ACM, CIE, and CCF.
\end{IEEEbiography}

\end{document}